\documentclass[a4paper]{article}

\usepackage{authblk}
\usepackage{amsfonts}
\usepackage{amsmath}
\usepackage{algorithm}
\usepackage{algorithmic}
\usepackage{url}
\usepackage{hyperref}
\usepackage{microtype}
\usepackage{subfig}
\usepackage{tikz}
\usetikzlibrary{positioning}
\usetikzlibrary{decorations.pathreplacing}

\newcommand{\N}{\mathbb{N}}
\newcommand{\F}{\mathbb{F}}

\providecommand{\keywords}[1]{\textbf{\textit{Keywords }} #1}

\begin{document}

\title{Tip the Balance: Improving Exploration of Balanced Crossover Operators by Adaptive Bias}

\author[1]{Luca Manzoni}
\author[2]{Luca Mariot}
\author[3]{Eva Tuba}

\affil[1]{{\normalsize Dipartimento di Matematica e Geoscienze, Università degli Studi di Trieste, Via Valerio 12/1, 34127 Trieste, Italy} \\
	
	{\small \texttt{lmanzoni@units.it}}}

\affil[2]{{\normalsize Cyber Security Research Group, Delft University of Technology,, Mekelweg 2, Delft, The Netherlands} \\
	
	{\small \texttt{l.mariot@tudelft.nl}}}

\affil[3]{{\normalsize Faculty of Informatics and Computing, Singidunum University Danijelova 32, 11000 Belgrade, Serbia} \\
	
	{\small \texttt{etuba@ieee.org}}}

\maketitle

\begin{abstract}
The use of balanced crossover operators in Genetic Algorithms (GA) ensures that the binary strings generated as offsprings have the same Hamming weight of the parents, a constraint which is sought in certain discrete optimization problems. Although this method reduces the size of the search space, the resulting fitness landscape often becomes more difficult for the GA to explore and to discover optimal solutions. This issue has been studied in this paper by applying an adaptive bias strategy to a counter-based crossover operator that introduces unbalancedness in the offspring with a certain probability, which is decreased throughout the evolutionary process. Experiments show that improving the exploration of the search space with this adaptive bias strategy is beneficial for the GA performances in terms of the number of optimal solutions found for the balanced nonlinear Boolean functions problem.

\end{abstract}


\section{Introduction}
\label{sec:introduction}
When dealing with certain optimization problems coming from the area of combinatorics, cryptography, and coding theory, it is often required that the candidate solutions satisfy a balancedness constraint, or more precisely that the bitstrings representing them have a specified \emph{Hamming weight} (i.e., a fixed number of ones). In the context of \emph{Genetic Algorithms} (GA), however, traditional variation operators such as one-point crossover and flip mutation which are used to explore the search space cannot guarantee that the generated solutions preserve the required Hamming weight~\cite{millan1998}.

In general, one can envision two main approaches to cope with this issue. The first one is to consider the balancedness constraint as a further property to be optimized in addition to those already taken into account. In a single-objective optimization setting, this usually amounts to adding a \emph{penalty factor} in the fitness function which punishes deviation from the desired Hamming weight. This approach has the advantage of yielding a GA with deterministic running time. Given enough fitness evaluations, the population always converges to the desired Hamming weight, since the penalty factor is a rather easy optimization objective to meet. However, in this way, the GA explores a search space that is much larger than the admissible set of balanced bitstrings, and this usually results in sub-optimal solutions concerning the other optimization objectives. The second method consists in designing \emph{ad-hoc variation operators}, which ensures that the new solutions generated throughout the optimization process all have the required number of ones. In this way, the search space is greatly reduced, since the GA is constrained to explore only the space of feasible solutions.

As far as our knowledge goes, the first attempt at designing a balanced crossover operator for GA dates back to the work of Millan et al.~\cite{millan1998}. There, the authors considered the problem of evolving highly nonlinear balanced Boolean functions for cryptographic applications, and they employed a counter-based crossover operator to ensure that the truth table of the offspring function is composed of an equal number of zeros and ones. Later, Chen et al.~\cite{chen2006,chen2009} proposed a combination genetic algorithm for evolving investment portfolios, where the underlying crossover operator preserved the Hamming weight of the vector specifying which assets to invest in. Meinl and Berthold~\cite{meinl2009} investigated a balanced two-point and a uniform crossover operator for the $k$-subset selection problem, which has relevance for virtual screening of molecules in drug design. More recently, Mariot and Leporati~\cite{mariot2015} modified the aforementioned counter-based crossover of Millan et al. to cope with three-valued strings, which were evolved as Walsh spectra of plateaued Boolean functions. The same approach as been extended in~\cite{mariot2017} to evolve balanced quaternary strings representing pairwise-balanced cellular automata rules, which were then used to build orthogonal Latin squares and in~\cite{mariot2018} to search for binary orthogonal arrays.

The approach of designing ad-hoc variation operators has been recently studied by the authors in~\cite{manzoni2020}, where three different \emph{balanced crossover operators} have been investigated in the context of three optimization problems related to cryptography and combinatorial designs. In particular, the authors assessed that the use of balanced crossover operators gives an advantage over one-point crossover for the considered problems. However, the results also showed that over large problem instances these balanced operators usually produce sub-optimal solutions for the optimization properties other than the balancedness constraint. The reason could lie in the fact that the reduced search space induced by the balanced operators has many isolated local optima where the GA gets stuck. 

In this paper, we investigate a hybrid approach where we allow a balanced crossover operator to produce \emph{partially unbalanced} candidate solutions with a certain probability. Similarly to the philosophy underlying \emph{simulated annealing}, the rationale is to accept in the early stages of the optimization process slightly worse solutions with respect to the balancedness constraint, to allow the GA to escape local optima and improve its exploration capabilities, and then to focus only on a region of the admissible solution set by decreasing the probability of introducing unbalancedness in the offspring. In particular, we modify the counter-based crossover operator originally proposed by Millan et al.~\cite{millan1998} and later considered in the experimental evaluation of~\cite{manzoni2020} by introducing an \emph{adaptive bias strategy}, where the probability of adding unbalancedness in the offspring is gradually decreased by using a \emph{geometric cooling mechanism} analogous to the one employed in simulated annealing.

We experimentally evaluate this adaptive bias strategy on the optimization problem of \emph{balanced nonlinear Boolean functions}, where the objective is to find a highly nonlinear Boolean function of $n$ variables whose truth table is composed by an equal number of ones and zeros. More precisely, we perform a parameter sweep for the unbalancedness probability and the cooling factor over the problem instance of Boolean functions of $n=7$ variables, showing that three parameter combinations for the adaptive bias strategy can produce optimal balanced solutions of nonlinearity $56$. Although the number of such solutions is still limited concerning the number of experimental runs performed, we remark that nonetheless, our results improve on the original counter-based crossover considered in~\cite{manzoni2020}, where the best solutions obtained have nonlinearity $54$. Thus, this adaptive bias strategy seems to represent a promising approach to be further investigated in the context of balanced crossover operators.

The rest of this paper is organized as follows. Section~\ref{sec:partially-unbalanced} recalls the details of the counter-based crossover operator described in~\cite{manzoni2020} and then describes our adaptive bias strategy to modify this crossover. Section~\ref{sec:experiments} presents the experimental evaluation of the adaptive bias strategy. Finally, Section~\ref{sec:conclusions} sums up the main contributions of the paper and points out some possible future directions of research on the topic.

\section{Partially Unbalanced Crossover}
\label{sec:partially-unbalanced}
In this section, we first delve into the details of the balanced crossover operators based on counters originally proposed in~\cite{millan1998}, whose performances with respect to one-point crossover have been later investigated in~\cite{manzoni2020}. We then introduce our adaptive bias strategy by modifying this crossover operator to allow the generation of partially unbalanced offspring with a specified probability.

\subsection{Counter-Based Balanced Crossover}
\label{subsec:cb-cross}
In what follows, we consider a bitstring of length $n \in \N$ as a vector of the $n$-dimensional vector space $\F_2^n$, where $\F_2=\{0,1\}$ is the finite field with two elements. Given $x \in \F_2^n$, the \emph{support} of $x$ is the set $supp(x) = \{i: 1\le i \le n, \ x_i \neq 0\}$ which specifies the positions of the elements set to $1$ in $x$. The \emph{Hamming weight} of $x$ is then defined as $w_H(x) = |supp(x)|$, i.e. the number of ones in $x$. The number of $n$-bit strings with a specified Hamming weight $1\le k \le n$ is $\binom{n}{k}$, since it is equal to the number of ways in which one can select a subset of $k$ elements from a set of cardinality $n$. This one-to-one relationship can be easily seen by identifying a bitstring $x \in \F_2^n$ of Hamming weight $k$ with the characteristic function of the corresponding subset of $k$ elements.

\begin{figure}[t]
\centering
\begin{tikzpicture}
      [->,auto,node distance=1.5cm, empt node/.style={font=\sffamily,inner
        sep=0pt,minimum size=0.6cm},
      rect0 node/.style={rectangle,draw,fill=olive,font=\sffamily\bfseries,minimum size=0.6cm, inner
        sep=0pt, outer sep=0pt},
      rect1 node/.style={rectangle,draw,fill=yellow,font=\sffamily\bfseries,minimum size=0.6cm, inner
        sep=0pt, outer sep=0pt},
      rect2 node/.style={rectangle,draw,fill=blue!50,font=\sffamily\bfseries,minimum size=0.6cm, inner
        sep=0pt, outer sep=0pt},]
      
      \node [empt node] (p)   {};
      \node [rect0 node] (p1) [right=0.7cm of p] {1};
      \node [rect0 node] (p0) [left=0cm of p1]  {0}; 
      \node [rect0 node] (p2) [right=0cm of p1] {0};
      \node [rect0 node] (p3) [right=0cm of p2] {1};
      \node [rect0 node] (p4) [right=0cm of p3] {0};
      \node [rect0 node] (p5) [right=0cm of p4] {1};
      \node [rect0 node] (p6) [right=0cm of p5] {1};
      \node [rect0 node] (p7) [right=0cm of p6] {0};

      \node [empt node] (b1) [left=0.1cm of p0] {$x$};

      \node [rect1 node] (q0) [below=0.8cm of p0] {1};
      \node [rect1 node] (q1) [right=0cm of q0] {0};
      \node [rect1 node] (q2) [right=0cm of q1] {0};
      \node [rect1 node] (q3) [right=0cm of q2] {0};
      \node [rect1 node] (q4) [right=0cm of q3] {1};
      \node [rect1 node] (q5) [right=0cm of q4] {0};
      \node [rect1 node] (q6) [right=0cm of q5] {1};
      \node [rect1 node] (q7) [right=0cm of q6] {1};

      \node [empt node] (b2) [left=0.1cm of q0] {$y$};

      \node [rect1 node] (c0) [below=0.8cm of q0] {1};
      \node [rect0 node] (c1) [right=0cm of c0] {1};
      \node [rect0 node] (c2) [right=0cm of c1] {0};
      \node [rect1 node] (c3) [right=0cm of c2] {0};
      \node [rect1 node] (c4) [right=0cm of c3] {1};
      \node [rect0 node] (c5) [right=0cm of c4] {1};
      \node [rect2 node] (c6) [right=0cm of c5] {0};
      \node [rect2 node] (c7) [right=0cm of c6] {0};

      \node [empt node] (b3) [left=0.1cm of c0] {$z$};

      \node [empt node] (f1) [below=0.4cm of q3.east] {\LARGE $\Downarrow$};
      \node [empt node] (f2) [right=0cm of f1] {\huge $\chi$};

      \node [empt node] (b4) [below=0.7cm of c1.east] {$cnt_1 = 4$};
      \node [empt node, minimum size=0cm] (e7) [below=0.4cm of c6.east] {};
      \node [empt node] (b5) [below=0.7cm of c6.east] {fill with $0$};

      \draw [>=stealth, thick] (c5.south) -- (b4.north);
      \draw [-,thick, decorate, decoration={brace,mirror,amplitude=3pt,raise=0.3cm}]
      (c6.west) -- (c7.east) node [midway,yshift=0.9cm] {};
      \draw [>=stealth, thick,inner sep=0pt] (b5.north) -- (e7);

    \end{tikzpicture}
    \caption{Example counter-based crossover applied over two $8$-bit strings with Hamming weight $w_H=4$.}
    \label{fig:cb-cross}
\end{figure}
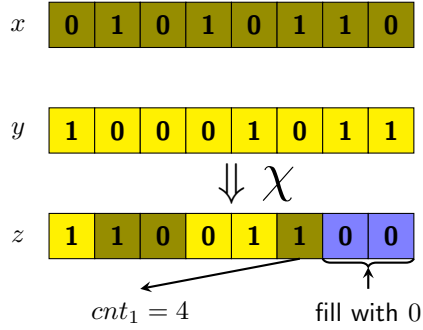

Given two bitstrings $x,y \in \F_2^n$ of Hamming weight $k$, the aim of a balanced crossover operator is to produce an offspring bitstring $z \in \F_2^n$ having the same weight $k$ as the parents. As mentioned in the previous section this can be accomplished by adopting several encoding for the candidate solutions, three of which have been explored in~\cite{manzoni2020}. Directly working on the bitstring representation is the most straightforward option, and this is the approach adopted in the \emph{counter-based crossover operator} originally proposed by Millan et al. in~\cite{millan1998}. The main idea underlying this operator is to build the offspring by copying bit by bit either from the first or the second parent by selecting them at random, and then use two \emph{counters} $cnt_0$ and $cnt_1$ to control respectively the multiplicities of $0$ and $1$. As soon as one of the two counters reaches its prescribed threshold (which is $n-k$ for $cnt_0$, and $k$ for $cnt_1$), the remaining positions of the offspring are filled with the complementary value to maintain the balancedness constraint. More precisely, the child chromosome $z \in \F_2^n$ is built from $x,y \in \F_2^n$ of Hamming weight $k$ as follows:
\begin{description}
\item[Initialization] Set both $cnt_0$ and $cnt_1$ to $0$
\item[Loop] For all positions $i \in \{1,\cdots,n\}$ of the child $z \in \F_2^n$, do one the following:
\begin{itemize}
\item If $cnt_0 = n-k$, set $z_i$ to $1$
\item If $cnt_1 = k$, set $z_i$ to $0$
\item If both $cnt_0 < n-k$ and $cnt_1 < k$, randomly copy with uniform probability either $x_i$ or $y_i$ in $z_i$, and update the relevant counter
\end{itemize}
\item[Return] $z$
\end{description}
From the high-level description above, it is easy to see that the Hamming weight of the produced offspring is always $k$, since if $n-k$ zeros are reached then $z$ is filled only with $1$, while if $k$ ones are copied then $z$ is completed only with zeros. Figure~\ref{fig:cb-cross} reports an example of this counter-based crossover operator applied to two bitstrings $x, y \in \F_2^8$ of length $8$ and Hamming weight $k=4$. The cells colored in green in the offspring $z$ correspond to the positions copied from the first parent $x$, while those in green are taken from $y$. Finally, the rightmost two cells in blue are those that are forced to be set to $0$, since the counter $cnt_1$ reaches the threshold $k=4$ at position $i=6$.

Looking at the way this crossover operator works, one might wonder whether it introduces a positional bias in the offspring bitstrings. In fact, by going from left to right the last bits are more likely to be set deterministically, once one of the two counters reaches its threshold. This issue has been investigated in~\cite{manzoni2020}, where the authors also considered a \emph{shuffling version} of this crossover operator where the offspring positions are randomly permuted before starting to copy from the parents. Interestingly enough, the results in that work showed not only that the shuffled version of the counter-based crossover operator is not beneficial, but in certain optimization problems (i.e. the search of binary orthogonal arrays addressed in~\cite{mariot2018}) worsens the GA performances. For this reason, in the rest of this work, we only consider the basic "left-to-right" version described above of the counter-based crossover operator.

\subsection{Adaptive Bias Strategy}
\label{subsec:adap-bias}
We now introduce the adaptive bias strategy which we used to modify the counter-based crossover described in the previous section, to allow a certain amount of unbalancedness in the generated offspring. The motivation behind this idea is that by allowing the GA to explore a slightly larger space of candidate solutions it could be easier to escape local optima than just by searching in the set of admissible solutions satisfying the required balancedness constraint.

The adaptive bias strategy comes into play in the counter-based crossover once one of the two counters for the control of the Hamming weight reaches its respective threshold. Suppose that the counter-based crossover has been applied over two bitstrings $x,y \in \F_2^n$ and that the threshold of $k=4$ has been achieved for the counter of ones $cnt_1$ at a certain position $i$. For the $n-i$ remaining positions of the offspring, instead of directly copying $0$, the crossover \emph{continues} to copy $1$ with a certain \emph{unbalancedness probability} $p \in [0,1]$. Thus, at each position $i < j \le n$ a random number $r \in [0,1]$ is drawn with uniform probability, and if $r$ is less than $p$ then $1$ is copied in the $j$-th position of the offspring, and the process is repeated for the next position. On the other hand, in the case where $r \ge p$ the value $0$ is copied in $z_j$, and \emph{all remaining positions} are set deterministically to $0$. In this way, the probability of obtaining an offspring bitstring composed of $k+n-i$ ones (which would result in a high deviation from the desired Hamming weight $k$) is $p^{n-i}$. Symmetrically, the same process is applied if the threshold of $n-k$ zeros is first reached by $cnt_0$ at position $i$, by continuing to copy with probability $p$ the value $0$ in the positions $i < j \le n$ of the offspring $z$, and filling the remaining ones with value $1$ once the sampled random number $r$ is greater than or equal to $p$. The probability of generating an offspring with $2n-k-i$ zeros is therefore again $p^{n-i}$.

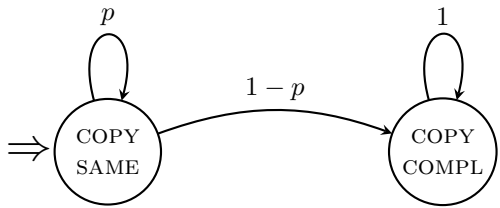
\begin{figure}[t]
\centering
    \begin{tikzpicture}
    [->,auto,node distance=1.5cm, every loop/.style={min distance=12mm},
           empt node/.style={font=\sffamily,inner sep=1pt,outer sep=0pt},
           circ node/.style={circle,thick,draw,font=\sffamily\bfseries,minimum size=1.4cm, inner sep=2pt, outer sep=0pt}]
    
           % Nodes
           \node [circ node] (q0) [align=center] {{\sc copy}\\{\sc same}};
           \node [empt node] (e0) [left=0cm of q0] {\LARGE $\Rightarrow$};
           \node [circ node] (q1) [right=3cm of q0, align=center] {{\sc copy}\\{\sc compl}};
    
           % Edges
           \draw [->, thick, shorten >=0pt,shorten <=0pt,>=stealth] (q0) 
           edge[bend left=20] node (f5) [above]{$1-p$} (q1);
           \draw[->, thick, shorten >=0pt,shorten <=0pt,>=stealth] (q0) edge[loop above] node (f3) [above]{$p$} ();
           \draw[->, thick, shorten >=0pt,shorten <=0pt,>=stealth] (q1) edge[loop above] node (f4) [above]{$1$} ();
    \end{tikzpicture}
    \caption{Finite state machine representation of the adaptive bias strategy.}
    \label{fig:fsa}
\end{figure}

Figure~\ref{fig:fsa} depicts the finite state diagram representing the technique described above.
In particular, the initial state is {\sc copy-same}, in which the same value associated to the counter that reached its threshold is still copied, thereby introducing unbalancedness in the offspring. The control stays in state {\sc copy-same} with probability $p$, while it changes to {\sc copy-compl} with probability $1-p$. After reaching {\sc copy-compl}, the crossover continues to copy the complementary bit value with probability $1$.

Algorithm~\ref{alg:cntcr} reports the pseudocode of our modified version of the counter-based crossover described in~\cite{manzoni2020}, which includes the adaptive bias strategy described above. The input to this algorithm are the two parent bitstrings $x,y \in \F_2^n$, their length $n$ and Hamming weight $k$, and the unbalancedness probability $p$. 
\begin{algorithm}[t]
\floatname{algorithm}{Algorithm}
\caption{{\sc Counter-Cross-Unbal}$(x, y, n, k, p)$}
\label{alg:cntcr}
\begin{algorithmic}[5]
\STATE $cnt_0$ := $0$; $cnt_1$ := $0$; $z$ := $0^n$; $i=0$;
\WHILE{$cnt_0 < n-k$ AND $cnt_1 < k$}
    \STATE $z_i$ := {\sc Random-Select}($x_i$, $y_i$)
    \IF{($z_i$ = $1$)}
        \STATE $cnt_1$ := $cnt_1+1$
    \ELSE
        \STATE $cnt_0$ := $cnt_0+1$
    \ENDIF
    \STATE $i$ := $i+1$
\ENDWHILE
\IF{$cnt_0 = n-k$}
    \STATE $val$ := $0$
    \ELSE
    \STATE $val$ := $1$
\ENDIF
\STATE $same$ := $TRUE$
\WHILE{$i<n$}
    \IF{$same$ = $TRUE$}
        \STATE $r$ := {\sc Random}$(0,1)$
        \IF{$r<p$}
            \STATE $z_i$ := $val$
        \ELSE
            \STATE $z_i$ := $val \oplus 1$
            \STATE $same$ := $FALSE$
        \ENDIF
        \ELSE
            \STATE $z_i$ := $val \oplus 1$
    \ENDIF
    \STATE $i$ := $i+1$
\ENDWHILE
\STATE return $z$        
\end{algorithmic}
\end{algorithm}
The first $10$ lines of Algorithm~\ref{alg:cntcr} implements the basic counter-based crossover before one of the counters reaches its threshold. Thus, each position $i$ of $z$ is determined by randomly copying either the value $x_i$ or $y_i$ with uniform probability, which is performed at line 3 by the subroutine {\sc Random-Select()}. Then, the relevant counter is increased depending on the copied value at lines 4--8. The {\bfseries if-else} block at lines 11--15 determines which is the bit value $val$ whose counter reached the corresponding threshold. Then, in the {\bfseries while} loop at lines 17--30 the offspring $z$ is completed with the adaptive bias strategy. In particular, the decision whether the same bit value $val$ or its complement $val \oplus 1$ must be copied in position $i$ is controlled by the flag $same$, which is set to false as soon as the random number $r$ sampled at line 19 is greater than or equal to the unbalancedness probability $p$. Once all positions have been filled, the algorithm finally returns the offspring $z$.

The use of the modified crossover operator described in Algorithm~\ref{alg:cntcr} allows generating bitstrings where the bias of the Hamming weight is related to the unbalancedness probability $p$. This parameter $p$ must be in turn controlled to drive the evolutionary process towards solutions whose number of ones increasingly approaches the target weight. Intuitively, the goal is to favor \emph{exploration} of the search space of slightly unbalanced bitstrings in the early first stages of the optimization process, and then to focus on the \emph{exploitation} of a subset of the space of bitstrings with the desired Hamming weight. To this end, we adopted a \emph{discount mechanism} of the unbalancedness mechanism inspired by the \emph{geometric cooling schedule} of simulated annealing~\cite{clark2004}. In particular, after a certain number $m$ of fitness evaluations, the unbalancedness probability is
updated as follows:
\begin{equation}
\label{eq:update}
p \gets \alpha \cdot p \enspace ,
\end{equation}
where $\alpha \in (0,1)$ is a \emph{cooling factor} analogous to that used in simulated annealing. In this way, the unbalancedness probability decreases exponentially, thus making the generation of unbalanced solutions more and more unlikely as the GA optimization process proceeds. In particular, if $p_0 \in (0,1)$ is the initial unbalancedness probability set at the beginning of the GA, the resulting probability after $t \in \N$ updates is $p(t) = \alpha^t \cdot p_0 \enspace$.

Further, to foster the selection of solutions with a Hamming weight close to the required one, we also adopted a \emph{penalty factor} at the fitness function level. Formally, given a bitstring $x \in \F_2^n$, this penalty factor is simply defined as the absolute value of the difference between the Hamming weight of $x$ and the target weight $k$, i.e. $pen(x) = |w_H(x) - k|$. In case of maximization problems, $pen(x)$ is subtracted from the fitness value of $x$, while for minimization problems it is added. In our experiments, we investigated two versions of the penalty factor: in the first one, the full value of the penalty factor is taken into account throughout the whole optimization process, while in the second one it is weighted with the complementary value of the unbalancedness probability $p$, that is
\begin{equation}
    \label{eq:wpen}
    w_{pen}(x) = (1-p)\cdot pen(x) \enspace . 
\end{equation}
The motivation underlying the use of this \emph{adaptive penalty factor} $w_{pen}(x)$ as defined in Equation~\ref{eq:wpen} is to discount the Hamming weight of the solutions produced in the early stages of the evolutionary process when computing their fitness values, thereby favoring exploration of the search space. Subsequently, as the unbalancedness probability $p$ decreases, $w_{pen}$ approaches the full value of the penalty factor, thus shifting the focus towards the selection of solutions with the desired Hamming weight.

\section{Experiments}
\label{sec:experiments}
In this section, we present the experimental evaluation that we performed on our adaptive bias strategy applied to the counter-based crossover operator. We first briefly introduce the optimization problem considered in the experiments, namely maximizing the nonlinearity of balanced Boolean functions. We then describe the experimental settings and parameters adopted for our tests, and finally, we present the obtained results.

\subsection{Balanced Nonlinear Boolean Functions}
\label{subsec:test-prob}
As a test problem, we considered the optimization of nonlinearity in balanced \emph{Boolean functions}, which comes from the area of cryptography and coding theory, and that has been tackled in several works with evolutionary algorithms (see e.g.~\cite{millan1998,mariot2015,picek2016}). In particular, this is one of the three problems considered in~\cite{manzoni2020} to investigate the performances of balanced crossover operators in GA, including the counter-based crossover described in Section~\ref{subsec:cb-cross}: thus, we used this problem to compare the results with those reported in~\cite{manzoni2020}.

We now briefly recall the basic notions about Boolean functions necessary to define our optimization problem of interest. For further details on this subject, the reader is referred to ~\cite{carlet2010}.

A \emph{Boolean function} of $n\in \N$ variables is a mapping of the form $f: \F_2^n \to \F_2$. The \emph{truth table} $\Omega_f$ of $f$ is the $2^n$-bit string which represents for each input vector $x \in \F_2^n$ the value of $f(x)$ in lexicographic order. For cryptographic applications, it is desirable that $f$ is \emph{balanced}, i.e. that its truth table is composed by an equal number of zeros and ones. In other words, the Hamming weight of $\Omega_f$ must be $2^{n-1}$. Another important cryptographic property is the \emph{nonlinearity} of $f$, which is the distance of $f$ from the set of all affine functions. This can be computed via the \emph{Walsh transform} of $f$, which is defined for all $a \in \F_2^n$ as:
\begin{equation}
    \label{eq:wt}
    W_f(a) = \sum_{x \in \F_2^n} (-1)^{f(x) \oplus a \cdot x} \enspace ,
\end{equation}
where $a \cdot x$ denotes the scalar product of $a$ and $x$. Then, the nonlinearity of $f$ is:
\begin{equation}
    \label{eq:nl}
    Nl(f) = 2^{n-1} - \frac{1}{2} \cdot \max_{a \in \F_2^n} \{|W_f(a)|\} \enspace .
\end{equation}
As a cryptographic criterion, the nonlinearity of $f$ should be as high as possible. Thus, the optimization problem requires finding a Boolean function of $n$ variables having highest possible nonlinearity. Given $f: \F_2^n \rightarrow \F_2$, we defined the following two fitness functions to be maximized, depending on whether the penalty factor is weighted or not:
\begin{align}
    \label{eq:fit1}
    fit_1(f) &= Nl(f) - pen(\Omega_f) = Nl(f) - |2^{n-1} - w_H(\Omega_f)| \enspace , \\
    \label{eq:fit2}
    fit_2(f) &= Nl(f) - wpen(\Omega_f) = Nl(f) - (1-p)\cdot |2^{n-1} - w_H(\Omega_f)| \enspace .
\end{align}

\subsection{Experimental Settings}
\label{subsec:exp-setting}
For the sake of comparison, we adopted an experimental setting closely matching the one used in~\cite{manzoni2020} in our experiments. As a test problem instance, we considered the space of Boolean functions of $n=7$ variables, since as noted in~\cite{manzoni2020}  for $n=6$ variables all balanced crossover operators converge quite easily on optimal balanced solutions of nonlinearity $26$. On the other hand, for $n=7$ variables only the map-of-ones crossover was able to produce one optimal balanced solution of nonlinearity $56$ over $50$ experimental runs, while all the other ones reached a maximum nonlinearity of $54$.

We employed a steady-state GA with a tournament selection operator, where at each iteration $t=3$ random individuals from the population are sampled. The best two individuals in the tournament are then selected for crossover and mutation, while the third one is replaced by the offspring produced by the two parents. For the crossover, we compared the basic counter-based operator and its modified version with our adaptive bias strategy. Concerning mutation, we used the swap-based mutation operator of~\cite{manzoni2020} with the same mutation probability $p_m=0.7$. The population is composed of $50$ individuals, and we stop the GA after $fit=1\,000\,000$ fitness evaluations. In particular, for our adaptive bias strategy, we updated the unbalancedness probability every $2\,000$ fitness evaluations, thus a total of $500$ updates took place during an evolutionary run. Finally, each experiment is repeated for $R=50$ independent runs.

\subsection{Results}
\label{subsec:results}
As far as we know, there are no studies in the literature suggesting what is the best amount of unbalance that can improve the performance of optimization algorithms for the search of highly nonlinear balanced Boolean functions. For this reason, we performed a parameter sweep over the two main values characterizing our adaptive bias strategy, namely the unbalancedness probability $p$ and the cooling factor $\alpha$. In particular, for $p$ we considered the values in the set $\{0.5, 0.6, 0.7, 0.8, 0.9\}$, while for $\alpha$ we adopted the values in $\{0.9, 0.95, 0.99\}$. In particular, for the latter parameter, we chose to focus on values close to $1$ in order to have a slow decrease in the unbalancedness probability, similarly to the works that used simulated annealing to search for cryptographic Boolean functions~\cite{clark2004,mariot2015}.

\begin{figure}[t]
    \subfloat[Full penalty factor\label{fig:heat1}]{
    \includegraphics[scale=0.26]{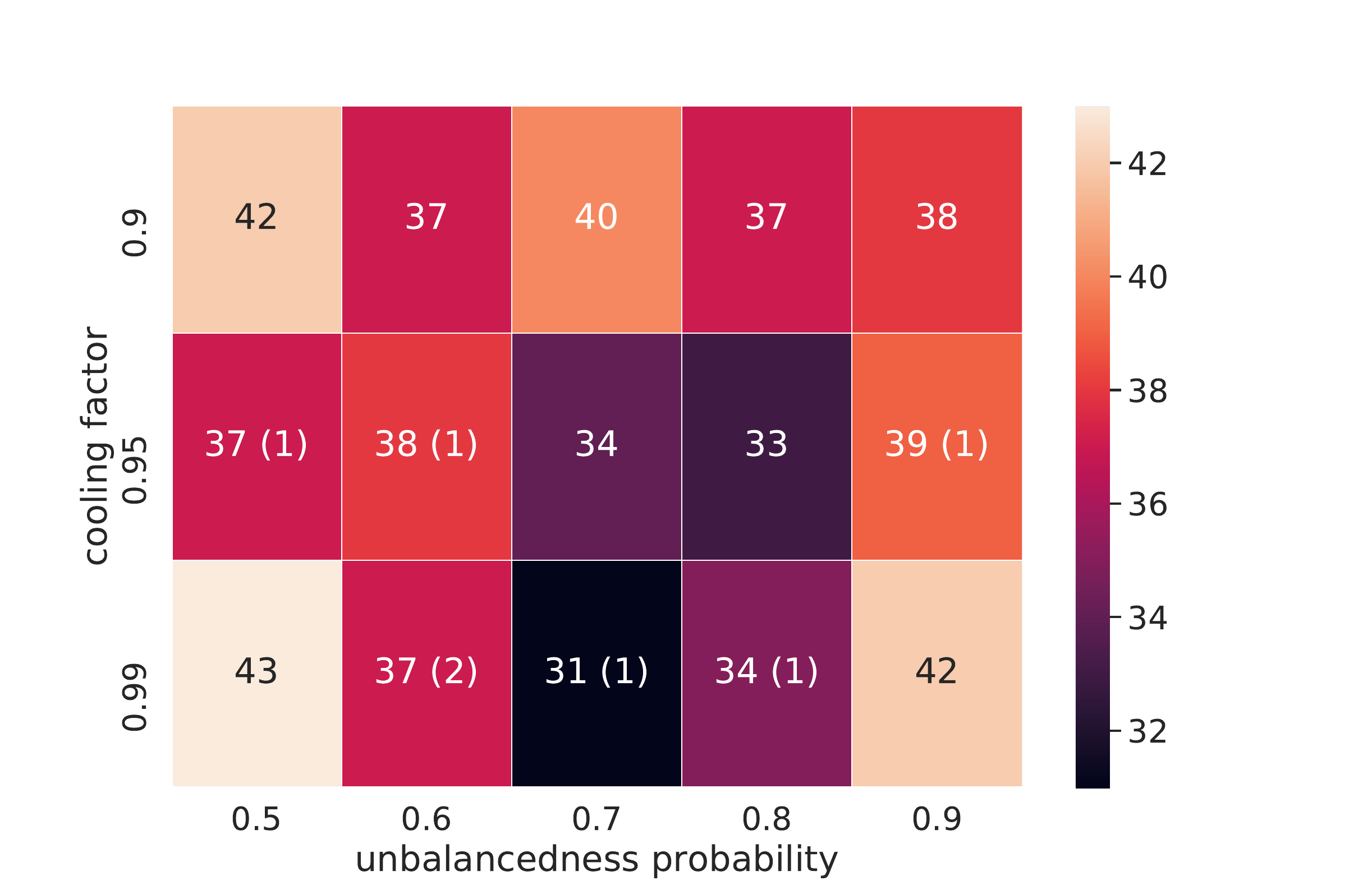}
    }
    \subfloat[Weighted penalty factor\label{fig:heat2}]{
    \includegraphics[scale=0.26]{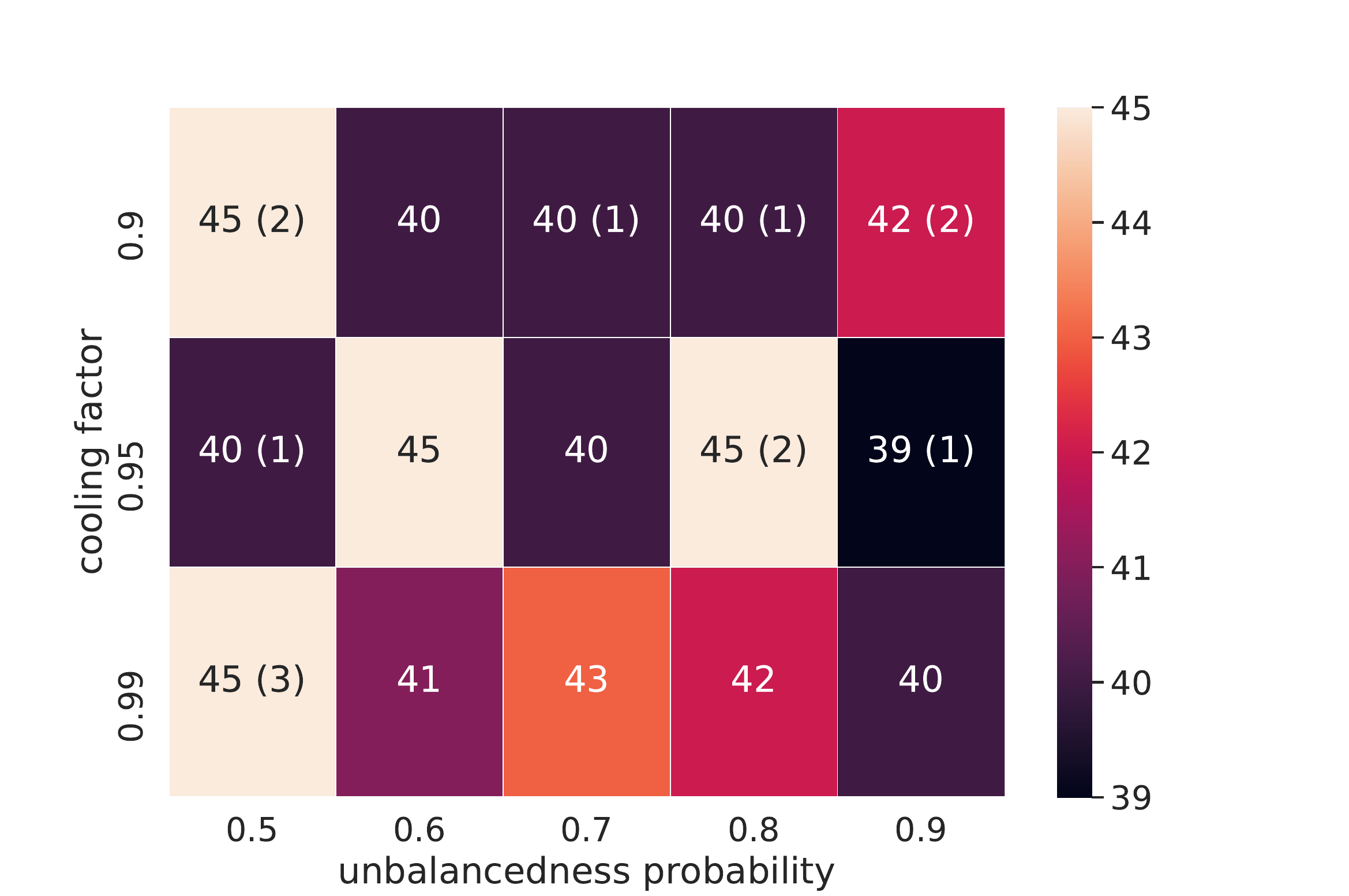}
    }
  \caption{Heatmaps for the parameter sweep over $p \in \{0.5, 0.6, 0.7, 0.8, 0.9\}$ and $\alpha \in \{0.9, 0.95, 0.99\}$ on the space of Boolean functions of $n=7$ variables.}
  \label{fig:ex-distance}
\end{figure}

Figures~\ref{fig:heat1} and ~\ref{fig:heat2} report the heatmaps of our parameter sweep experiments respectively for fitness function $fit_1$ (where the full penalty factor $pen(\cdot)$ is considered) and $fit_2$ (where the dynamically weighted penalty $wpen(\cdot)$ is used instead).
For each parameter combination $(p,\alpha)$, the numerical entry in the corresponding heatmap reports the number of best balanced solutions found by the GA over the $50$ experimental runs which reached a nonlinearity of at least $54$. The second number in parentheses, if present, indicates the number of solutions reaching nonlinearity $56$ (which is the maximum possible for balanced Boolean functions of $7$ variables). In general, it can be observed that using the dynamically weighted penalty factor $wpen$ improves the performances of the adaptive bias strategy, since all parameter combinations except one reach at least the $80\%$ of experimental runs where the best solution found has nonlinearity greater than or equal to $54$. The same applies also for the number of parameter combinations finding optimal solutions of nonlinearity $56$. For this reason, we considered only the results with the dynamically weighted penalty factor for our subsequent comparison.

\begin{figure}[t]
\centering
\includegraphics[scale=0.45]{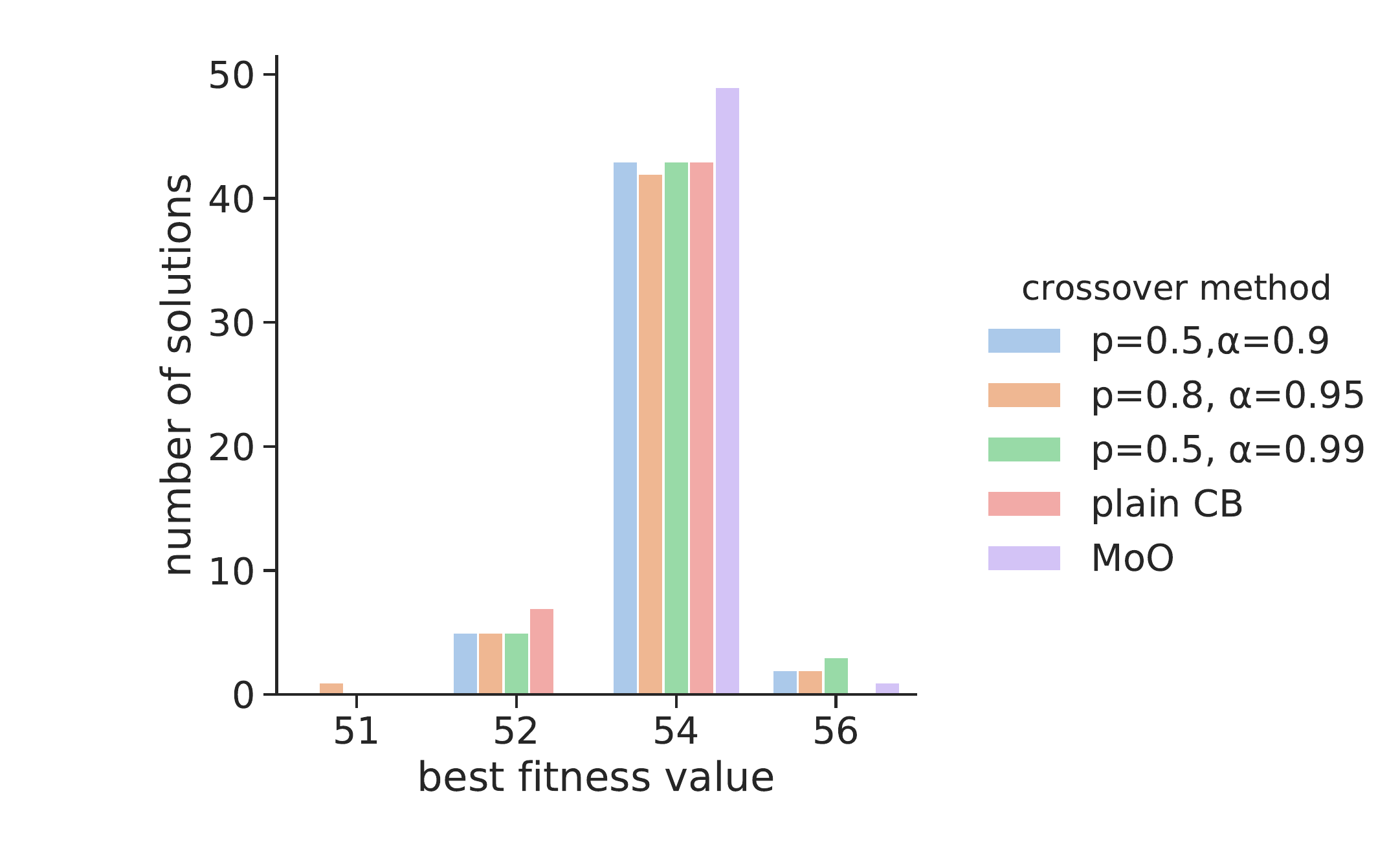}
\caption{Distribution of the fitness values for the five compared methods.}
\label{fig:hist}
\end{figure}

To compare our results, we selected the best parameter combinations for the adaptive bias strategy with the weighted penalty factor achieving a $90\%$ success rate. Here the success rate is defined as the number of runs where the best solutions have nonlinearity $54$, and where \emph{at least} one run produced an optimal solution of nonlinearity $56$. This selection resulted in the three parameter combinations $(p=0.5,\alpha=0.9)$, $(p=0.5,\alpha=0.99)$ and $(p=0.8,\alpha=0.95)$. We then compared their results with those achieved by the basic counter-based crossover and the map-of-ones crossover analyzed in~\cite{manzoni2020}. In particular, according to the results of~\cite{manzoni2020} the map-of-ones operator scored the best performance over the problem instance of $n=7$ variables. Notice that we tested these two crossover operators using the same experimental settings showed in Section~\ref{subsec:exp-setting}, which are the same as those adopted in~\cite{manzoni2020} except for the larger number of fitness evaluations considered ($1\, 000 \, 000$ instead of $500\,000$). Figure~\ref{fig:hist} depicts the distribution of the best fitness values achieved by the three selected parameter combinations of our adaptive bias strategy and by the basic counter-based and map-of-ones crossover (respectively denoted as plain CB and MoO).
It can be remarked from the plot that the three best combinations of our adaptive bias strategy can produce more optimal solutions of nonlinearity $56$ than both the basic counter-based crossover (which produced none) and the map-of-ones crossover (which produced only one out of $50$ experimental runs). In particular, the combination $(p=0.5,\alpha=0.99)$ produced $3$ optimal solutions.

\section{Conclusions}
\label{sec:conclusions}
In this paper, we proposed an adaptive bias strategy to perturb the behavior of a counter-based balanced crossover operator, to introduce a certain amount of unbalancedness in the offspring. The motivation behind this strategy is to improve the GA exploration of the search space in the first stages of the optimization process, to avoid early convergence to local optima. This is accomplished in the crossover operator by continuing to copy the binary allele that already reached the prescribed threshold with a specific unbalancedness probability $p$, while with probability $1-p$ the chromosome is completed with the complementary value. The unbalancedness probability is then exponentially decreased by multiplying it by a cooling factor of $\alpha$. We tested this approach on the problem of maximizing the nonlinearity of balanced Boolean functions and compared our results with the balanced crossovers analyzed in~\cite{manzoni2020}. Specifically, we focused on the space of Boolean functions of $n=7$ variables, and we performed a parameter sweep for tuning both the unbalancedness probability $p$ and the cooling factor $\alpha$. The results showed that using a dynamically weighted penalty factor in the fitness function where the weight is the complement of the unbalancedness probability allows generating more solutions with nonlinearity greater than or equal to $54$. Next, we compared the three best parameter combinations emerging from our parameter sweep with those achieved by the plain counter-based crossover and the map-of-one crossover defined in~\cite{manzoni2020}. The comparison showed that the three parameters combinations are able to produce slightly better results concerning the number of optimal balanced solutions with nonlinearity $56$.

Looking at the distribution plot of Figure~\ref{fig:hist}, a natural question is whether our adaptive bias strategy gives a significative advantage over the plain counter-based and the map-of-ones crossover. As a matter of fact, our best parameter combination can produce $3$ optimal balanced functions of nonlinearity $56$ out of $50$ experimental runs, while the map-of-ones generated only one of them. We remark however that the search of highly nonlinear balanced Boolean functions is known to be a difficult combinatorial optimization problem for GA~\cite{clark2004,mariot2015,picek2016,picek2016a}. In particular, while it is relatively easy to converge over an optimal balanced solution for $n=6$ variables, a steep increase in the difficulty of the problem can be observed on the $n=7$ problem instance~\cite{manzoni2019,manzoni2020}. Therefore, even a slight improvement in the number of optimal solutions produced over this problem instance is of interest, and our adaptive bias strategy seems to represent an interesting candidate for achieving this goal. This is further corroborated by the fact that multiple combinations considered in our parameter sweep were able to produce at least one optimal solution of nonlinearity $56$, which seems to rule out the possibility that our adaptive bias strategy found them by chance.

Clearly, further research in this direction is required to boost even more the performance of our adaptive bias strategy for the counter-based crossover operator. A first idea would be to perform some additional tuning of the unbalancedness probability and the cooling factor around the three combinations that yielded the best results in our parameter sweep, to assess how the GA performance changes by applying some small perturbances. It would also be interesting to investigate the fitness landscape over the space of Boolean functions of $n=7$ variables, to understand what amount of unbalancedness would be optimal for our adaptive bias strategy. One possibility in this respect would be to employ \emph{Local Optima Network} analysis, which the authors of~\cite{picek2019} performed for the case of vectorial Boolean functions. Finally, a third direction for future research is to investigate the performance of our adaptive bias strategy over other optimization problems that require balanced solutions, such as the bent functions problem or the orthogonal arrays problem considered in~\cite{manzoni2020} or the evolution of pairwise-balanced cellular automata rules to build orthogonal Latin squares studied in~\cite{mariot2017}.

\bibliographystyle{splncs04}
\bibliography{bibliography}

\end{document}